# A Novel Framework based on SVDD to Classify Water Saturation from Seismic Attributes

Soumi Chaki[1], Akhilesh Kumar Verma[2], Aurobinda Routray[1], William K. Mohanty[2], Mamata Jenamani[3]

[1]Department of Electrical Engineering, IIT Kharagpur, India
soumibesu2008@gmail.com
aroutray@ee.iitkgp.ernet.in

[2] Department of Geology and Geophysics, IIT Kharagpur, India
akhileshdelhi2007@gmail.com
wkmohanty@gg.iitkgp.ernet.in

[3]Department of Industrial and Systems Engineering, IIT Kharagpur, India
mj@iem.iitkgp.ernet.in

***Abstract*— Water saturation is an important property in reservoir engineering domain. Thus, satisfactory classification of water saturation from seismic attributes is beneficial for reservoir characterization. However, diverse and non-linear nature of subsurface attributes makes the classification task difficult. In this context, this paper proposes a generalized Support Vector Data Description (SVDD) based novel classification framework to classify water saturation into two classes (Class high and Class low) from three seismic attributes– seismic impedance, amplitude envelop, and seismic sweetness. G-metric means and program execution time are used to quantify the performance of the proposed framework along with established supervised classifiers. The documented results imply that the proposed framework is superior to existing classifiers. The present study is envisioned to contribute in further reservoir modeling.**

## I. Introduction

In the field of petroleum science, water saturation is a key parameter associated with reservoir system. It represents the fraction of formation water present in the pore space. Therefore, several studies are carried out to model or classify water saturation along with mutual effect analyses among water saturation and other petrophysical properties [1–3]. Modeling of reservoir properties are carried out using state-of-art machine learning algorithms such as Artificial Neural Networks (ANN), Fuzzy Logic (FL), Genetic Algorithm (GA), etc. Similarly, classification based approaches also facilitate reservoir characterization [4], [5]. Literature survey reveals that generally the supervised classifiers are selected over unsupervised counterparts due to the complexity associated with a given problem. However, the learning and performance of the supervised classifiers are dependent on the availability of a complete and representative training dataset. Therefore, supervised classifiers may not be able to deliver expected performance while dealing with an imbalanced dataset. In recent studies, the learning problems associated with an imbalanced dataset have gained attention from eminent researcher community for "real-world applications" [6–9]. In remote sensing fields, kernel based methods have emerged as a popular classification approach [10–12]. For example, a popular algorithm Support Vector Data Description (SVDD) is implemented in different problem areas due to its learning capacity irrespective of any prior knowledge on dataset [13–15]. Several classifiers based on discriminant [16–18], naive Bayes, support vector machine based classifier [19], [20], artificial neural network are used to solve different classification problems.

A SVDD based one-class classification framework is proposed in [4] to classify water saturation level from well logs. However, there are two limitations of the work reported in [4]. Firstly, seismic attributes are not included as predictor variables. Secondly, variation in water saturation level over the study area is not studied. These two shortcomings are addressed in the present study. In this paper, SVDD [13], [14] is used to design the classification framework to classify water saturation from seismic attributes. A dataset consists of multiple seismic attributes such as seismic impedance, amplitude envelop, and seismic sweetness along with well logs acquired from four wells in the area of interest is used in this study.

The contributions of the present study are as follows:

- a complete classification framework integrating seismic and well log signals
- blind prediction
- comparison with other classifiers
- water saturation level map over the area

This paper is organized as follows. In section II, description of the working dataset is given. Then, the theory of SVDD is briefly described in section III. Section IV describes the proposed classification framework and performance evaluators. Section V reports the experimental results. Finally, the paper is concluded with the discussion and future scope.

## II. Dataset Description

In this study, four sets of well logs acquired from a western onshore field of India are used. These four wells are to be referred as A, B, C, and D, henceforward. In a recent work, four well logs gamma ray content (GR), bulk density (RHOB), P-sonic (DT), and neutron porosity (NPHI) are used to classify a petrophysical property –water saturation[4] using a SVDD based novel classification framework. However, these logs are only available at some specific well locations inside the study area. To achieve an area map of water saturation level, seismic attributes are to be included as predictor variables instead of well logs. There are five seismic attributes acquired from the same study area such as seismic impedance, instantaneous amplitude, instantaneous frequency, amplitude envelop and seismic sweetness. However, only three of them are selected from all available attributes by appropriate relevant features selection algorithm. Seismic impedance, amplitude envelop, and seismic sweetness are selected over instantaneous amplitude and instantaneous frequency by Relief algorithm. In this study, we have classified water saturation from selected three seismic attributes by a SVDD based framework.

## III. Support Vector Data Description

Among the several methods available in the literature for the classification of large dataset into different classes, Support Vector Data Description (SVDD) is extensively used [13]. It is a useful method for different problems such as outlier detection, pattern recognition and classification, face recognition, etc. [4], [13–15].

SVDD is an extension of Support Vector Machines (SVMs). A boundary around a data set is constructed by SVDD algorithm. A hypersphere $F(R,a)$ defines the close boundary, where ' $a$ ' and ' $R$ 'are the center and radius of the hypersphere. It is stated that the volume of the hypersphere should be minimized for the data description[13–16]. The presence of outlier in the data is tested by defining a slacks variables $\varepsilon_i \geq 0$. The following error function is minimized to detect outliers,

$$F(R,a) = R^2 + C\sum_i \varepsilon_i \left\|x_i - a\right\|^2 \leq R^2 + \varepsilon_i \quad (1)$$

where,

$$\left\|x_i - a\right\|^2 \leq R^2 + \varepsilon_i, \ \text{ for all } i. \quad (2)$$

In addition, the SVDD function is represented as

$$L = \sum_i \alpha_i K(x_i, x_j) - \sum_{i,j} \alpha_i \alpha_i K(x_i, x_j)$$
$$\text{for all } \alpha_i : 0 \leq \alpha_i \leq C \quad (3)$$

Where, $K(x_i, x_j) = \phi(x_i).\phi(x_j)$ is a kernel function and it is used to have flexibility in the data description. This is an optimization problem which can be solved using Lagrange multipliers methods, i.e., by setting partial derivatives of $R$, $a$ and $x_i$ to zero. In this case, the constraints are $\sum_i \alpha_i = 1$ and $a = \sum_i \alpha_i x_i$ . Putting these values in (1) and (2), and then by minimizing $L$ we can determine the values of $\alpha_i$ [15]. In the present study, a Gaussian kernel is used to represent the function $K(x_i, x_j) = \phi(x_i).\phi(x_j)$ [16], [21]. The Gaussian kernel function is given by

$$K\left(x_i, x_j\right) = e^{-qx_i - x_j} \quad (4)$$

The objects with non-zero coefficients ($\alpha_i$) are called the support vectors, and only the support vectors are required in the description of the sphere. In order to determine whether a test point is within the sphere, the distance between test point and center of the sphere is determined. If this distance is smaller than the radius ($R$) then objects are accepted, i.e.,

$$R^2(x) = K(x,x) - 2\sum_i \alpha_i K(x_i, x) + \sum_{i,j} \alpha_i \alpha_j K(x_i, x_j) \quad (5)$$

In other words, we take the radius of the circle $R$ to be the maximum of values $R(x)$ for the support vectors. Hence, data points lying outside the circle of radius R are considered to be outliers. It is noted that the application of two-class classifier is helpful compared to its one-class counterpart while working with an imbalanced dataset [4], [22], [23].

## IV. Proposed Classification Framework

Literature studies reveal that SVDD along with other kernel based algorithms have emerged as efficient means to classify a property using an imbalanced dataset in various domains e.g. hyperspectral image processing, reservoir characterization, outlier detection, document classification etc. This paper proposes a novel classification framework to classify water saturation from seismic attributes using an imbalanced geological dataset. There are four steps included in the workflow namely– A) data preparation, B) preliminary analysis, C) training and testing, D) volumetric classification and visualization of water saturation level map as demonstrated in Fig. 1. The steps in the proposed framework are designed by modifying the published work in [4]and briefly described in this section.

### A. *Data Preparation*

Seismic attributes along with four sets of well log data are used in this study. As shown in the figure (Fig. 1), the procedure is started with data acquisition and integration of seismic and borehole data. First, the well logs are converted into time domain from depth domain using time-depth relationships available at specific well locations. Then, seismic attributes at specific four well locations are extracted from seismic volume. It is found that the sampling intervals of these dataset (seismic and well logs) are different. For example, the seismic patterns are sampled at an interval of two milliseconds, whereas the sampling interval of well logs is 0.15 milliseconds. Hence, we interpolate the band limited seismic signals at 0.15 milliseconds sampling interval corresponds to that of the well logs. Thus, two different domain signals are combined to create a dataset to be used in the given classification problem.

### B. *Preliminary Analysis*

The performance of classifiers is dependent on selection of relevant features. First, a number of "candidate features" are extracted from raw dataset. Then, different algorithms i.e. mutual information, Relief algorithm [24], and its variants are used to identify relevant features among available features before starting to train the classifier. In this paper, Relief algorithm selects statistically relevant features from a noisy dataset. Inclusion of unnecessary inputs in model elongates training time along with increase in the model complexity. In contrary, application of relevant features as predictor variables enhances the generalization capability of a model [4].

Then, the water saturation is classified into two classes, namely- Class high and Class low using a user defined threshold. The selection of threshold level is governed by two constraints [4]. First, a threshold level is selected such that the Class high samples are as close to maximum water saturation value and the Class low patterns remain close to minimum water saturation value. Next, training time of SVDD classifier plays a role in threshold selection. More importantly, selected threshold level is verified by an expert geophysicist. Similar work has been demonstrated in a recent paper [4]. However, the work carried out in [4]is associated with classification of water saturation from well logs. The drawback of [4]is that the water saturation levels are known only at specific well locations. In this paper, we have overcome this limitation by using seismic attributes as predictor variables. Therefore, an area map can be produced identifying high and low water saturation levels.

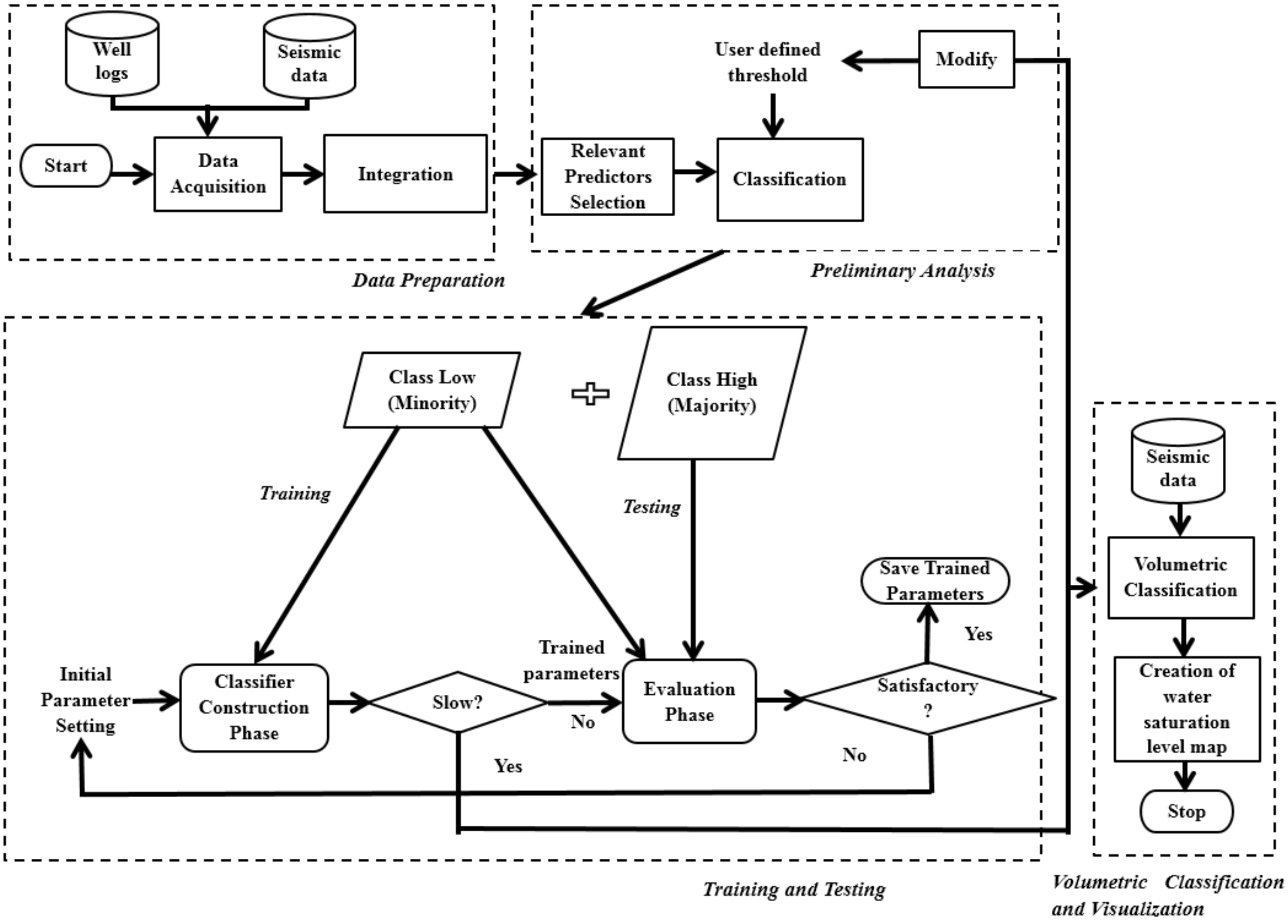

Fig. 1. Proposed classification framework

## C. *Training and Testing*

The lower part of Fig. 1 represents the training and testing steps associated with the classifier. For the working dataset, the number of available samples belong to Class high is significantly large which in turn makes it majority class. Conversely, Class low is minority class due to presence of small amount of samples belonging to this category in the working dataset. The division of training and testing pattern is carried out as in [4]. The minority class (Class low) patterns belong to integrated dataset of three wells are used to train the classifier. The tuned classifier parameters are validated using the Class low patterns of test well and the combined majority class (Class high) samples of all the wells.

The input attributes (seismic impedance, amplitude envelop, and seismic sweetness) of training patterns are used to construct the SVDD hypersphere. Classification accuracy of SVDD is improved by adjusting multiple parameters such as the kernel function and associated parameters, and radius of the hypersphere $C$. We have experimented with different kernel functions such as Gaussian, higher order polynomial (2–10), radial basis function, and exponential radial basis function along with associated kernel parameters with $C$ values varying from 0 to 1. The task of the classifier is to minimize the Lagrangian function by constrained optimization as mentioned earlier in Section III. The data samples are categorized into three categories: true data (inside the hypersphere), outliers (outside the hypersphere), and support vectors (at the hypersphere periphery) by this optimization. As in [4], the support vectors are encompassed in the outlier category. The tuned parameters are tested using the majority class samples.

To establish the proposed framework over existing classifier algorithms (e.g. ANN, and support vector machine based classifier), a comparison has been carried out. In all cases, the predictor attributes, and performance evaluators are same as the proposed framework. The division of training-testing samples and associated classification parameters are varied depending on respective classifiers.

The performance of the proposed framework is quantified using g-metric means [4], [25] and program execution time. G-metric means is associated with the accuracy of both positive and negative classes and often used in case of imbalanced dataset.

### *D. Volumetric Classification and Visualization*

The trained parameters which yield acceptable results in the blind testing are saved. Then, the water saturation level in the study area can be estimated from seismic attributes. The saved SVDD parameters classify the water saturation level in Class high or Class low at any location in the study area using seismic attributes of the area. After the classification over the area, the variation of water saturation level is visualized at any selected part of the study area.

## V. Experimental Results

The research work carried out in this study are performed on a 64 bit MATLAB platform installed on a Intel(R) Core(TM) i5CPU @3.20 GHz workstation having 16 GB RAM. The following sections describe the experimental results achieved in every steps of the proposed framework.

### *A. Dataset Preparation*

The proposed framework starts with dataset preparation. The integration of seismic attributes and petrophysical properties (e.g. water saturation) is an important step. First, seismic attributes are extracted at the available four well locations from seismic dataset. The well logs are represented in depth domain, whereas, seismic signals are recorded in time domain. The water saturation log is converted to time domain from depth domain using time-depth relationships available at well locations.

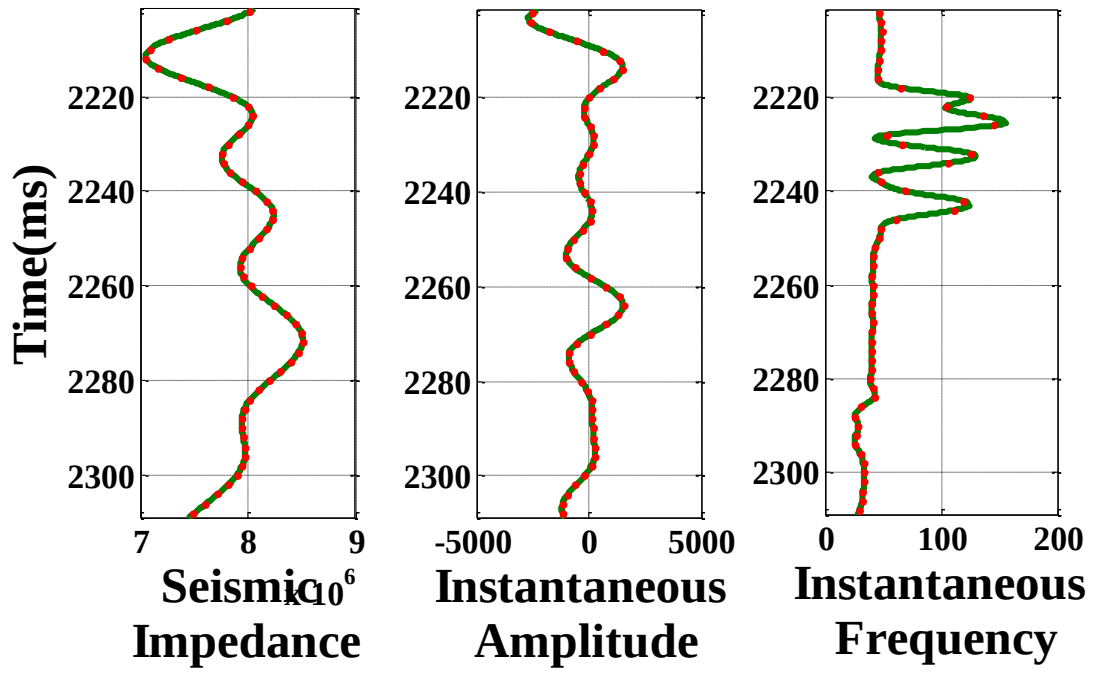

(a) (b) (c)

Fig. 2: Plots of (a) seismic impedance, (b) instantaneous amplitude, and (c) instantaneous frequency along time (ms) for well A

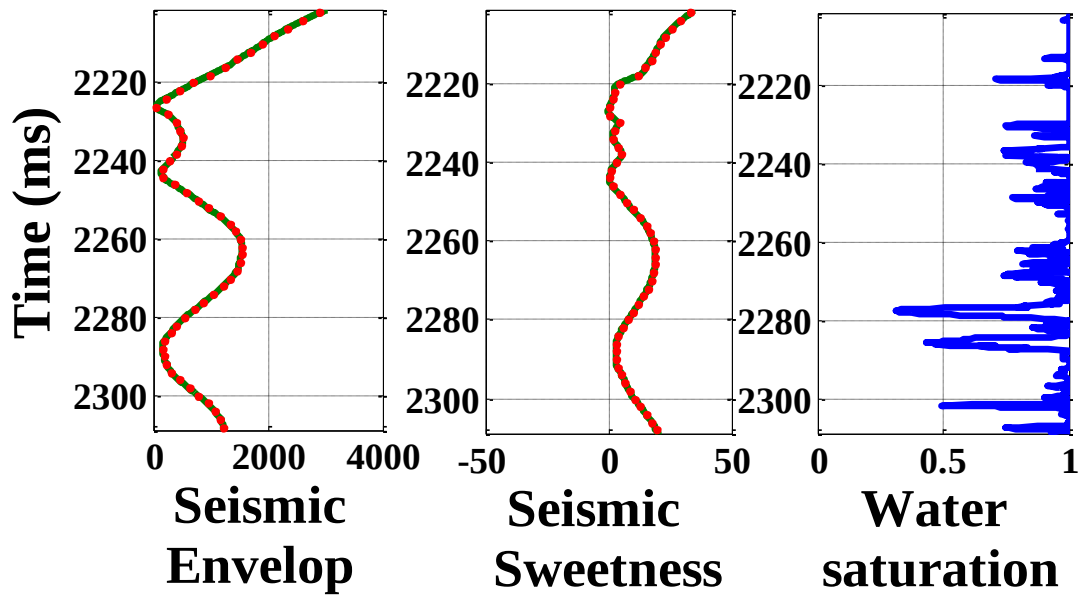

(a) (b) (c)

Fig. 3: Plots of (a) seismic amplitude envelop, (b) seismic sweetness, and (c) water saturation along time (ms) for well A

After that, the difference in sampling interval of seismic and well logs are observed. So, seismic signals are interpolated using spline interpolation at 0.15 milliseconds sampling interval pertaining to the well logs to integrate the seismic and water saturation signals at four well locations to form the master dataset. Figs. 2–3 represent available five seismic attributes- (2(a)) seismic impedance, (2(b)) instantaneous amplitude, (2(c)) instantaneous frequency, (3(a)) seismic amplitude envelop, (3(b)) seismic sweetness and (3(c)) water saturation along the well A. The red dots on the seismic attributes represent original values at time interval of two milliseconds and the green curves represent reconstructed signals along the time interval of well log data. The blue curve in Fig. 3(c) represents water saturation along the well A. It can be observed that water saturation distribution is biased towards maximum water saturation value (i.e. one).

### *B. Preliminary Analysis*

We have started with five seismic attributes extracted from seismic volume at four well locations and integrated with water saturation logs. To train the classifier using relevant features in order to avoid prolonged learning time, three relevant features are selected from available five "candidate attributes" using Relief algorithm. The result of Relief algorithm is represented in Fig. 4. Fig. 4 reveals that seismic impedance, seismic amplitude envelop, and seismic sweetness are more relevant features with respect to water saturation in terms of predictor importance weight compared to instantaneous amplitude and instantaneous frequency.

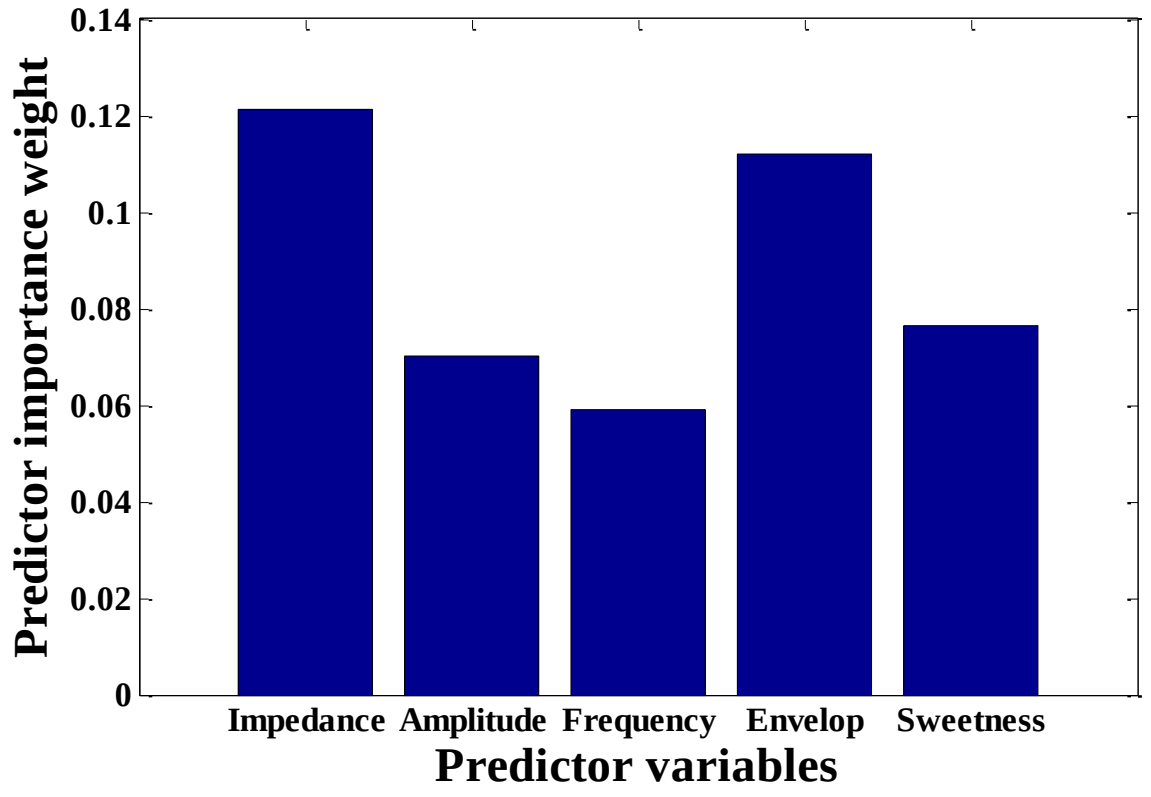

Fig. 4: Selection of relevant input attributes using Relief algorithm

The inclusion of the relevant predictor attributes selection step in the methodology is carried out to elude the possibility of unnecessary increase in the complexity of the classifier. The program execution time also increases with the increase in number of predictor variables.

The next job is to initialize the user define threshold to classify water saturation levels into two classes: Class low and Class high. We selected the initial threshold level as 0.7 as in [4].

### C. *Training and Testing*

The training and testing of the classifier are carried out following the work done in[4]. The initial kernel function and values of associated parameters are selected intuitively. Then, depending upon the improvement of classification result, these variables are empirically modified. For example, the blind testing results of well A as documented in this study is achieved after training the proposed SVDD classifier with a Gaussian kernel having 3.0 as width parameter and $C$ value of 0.005 with minority class patterns belong to remaining three wells (Well B, C, and D).

The proposed classifier is compared with SVM, and ANN based classifiers. These classifiers are optimized with appropriate parameter values related to respective algorithms. In case of ANN, the number of hidden layer neurons are selected in a way such that total number of weights and biases are at least fifteenth time less than the number of the available training samples. Thus, the possibility of over fitting of the ANN is avoided. The predictor variables are same (seismic impedance, seismic amplitude envelop, and seismic sweetness) as that of the proposed framework. The difference lies in the creation of training and testing data set. For these classifiers, the learning is carried out using the integrated dataset of three wells. The samples corresponding to the remaining fourth well are used to test the trained classifiers. Thus, majority and minority class components are collectively used to train the network instead of using only minority class patterns. The training and testing sets are different in case of ANN, SVM based classifiers from the proposed framework. However, the training and testing cases are mutually exclusive in each cases. Moreover, the testing set used for ANN, and SVM based classifiers are a subset of that of the testing set pertaining to the proposed framework. The proposed framework is trained using only the minority class patterns of the training set. Therefore, the results attained in terms of g-metric mean and program execution time are unbiasedly achieve produce better results than the existing algorithms.

Table I and Table II represent the comparison results of proposed framework with other three classifiers in terms of g-metric mean and program execution time (in seconds) respectively. It can be observed from Table I that the g-metric mean values in case of ANN based classifier are very poor. Then, the blind testing performance improves while using kernel based algorithm SVM based classifier. Finally, our framework has yield better performance compared to both– ANN and SVM based classifiers in reduced time. As number of patterns belongs to minority class is insignificant compared to that of the majority class, hence, trained classifiers are able to detect the majority class testing patterns correctly. However, the minority class test patterns are also wrongfully classified in Class high (majority class). Hence, g-metric mean is poor. On the other hand, our framework is based on one class classification. Therefore, it is able to detect minority class patterns in testing dataset yielding better g-metric means.

TABLE I: PERFORMANCE COMPARISON OF CLASSIFIERS IN TERMS OF G-METRIC MEAN

| Well Name | Value of g-metric mean | | |
|---|---|---|---|
| | *Artificial Neural Network based Classifier (ANN)* | *SVM* | *Proposed Workflow (SVDD)* |
| A | 0.28 | 0.48 | 0.72 |
| B | 0.26 | 0.65 | 0.74 |
| C | 0.34 | 0.55 | 0.69 |
| D | 0.20 | 0.62 | 0.65 |
| Average Performance | 0.27 | 0.57 | 0.7 |

TABLE II: PERFORMANCE COMPARISON OF CLASSIFIERS IN TERMS OF PROGRAM EXECUTION TIME (IN SECONDS)

| Well Name | Value of g-metric mean | | |
|---|---|---|---|
| | *Artificial Neural Network based Classifier (ANN)* | *SVM* | *Proposed Workflow (SVDD)* |
| A | 26.834 | 16.74 | 12.37 |
| B | 20.238 | 18.84 | 14.2 |
| C | 21.523 | 15.14 | 12.25 |
| D | 22.839 | 14.64 | 13.57 |
| Average Performance | 22.8585 | 16.34 | 13.09 |

The results in Table I and Table II are pictorially represented in Fig. 5 and Fig. 6 respectively. Fig. 5 and Fig. 6 reveal that the proposed framework has attained better performance compared to other classifiers in reduce time.

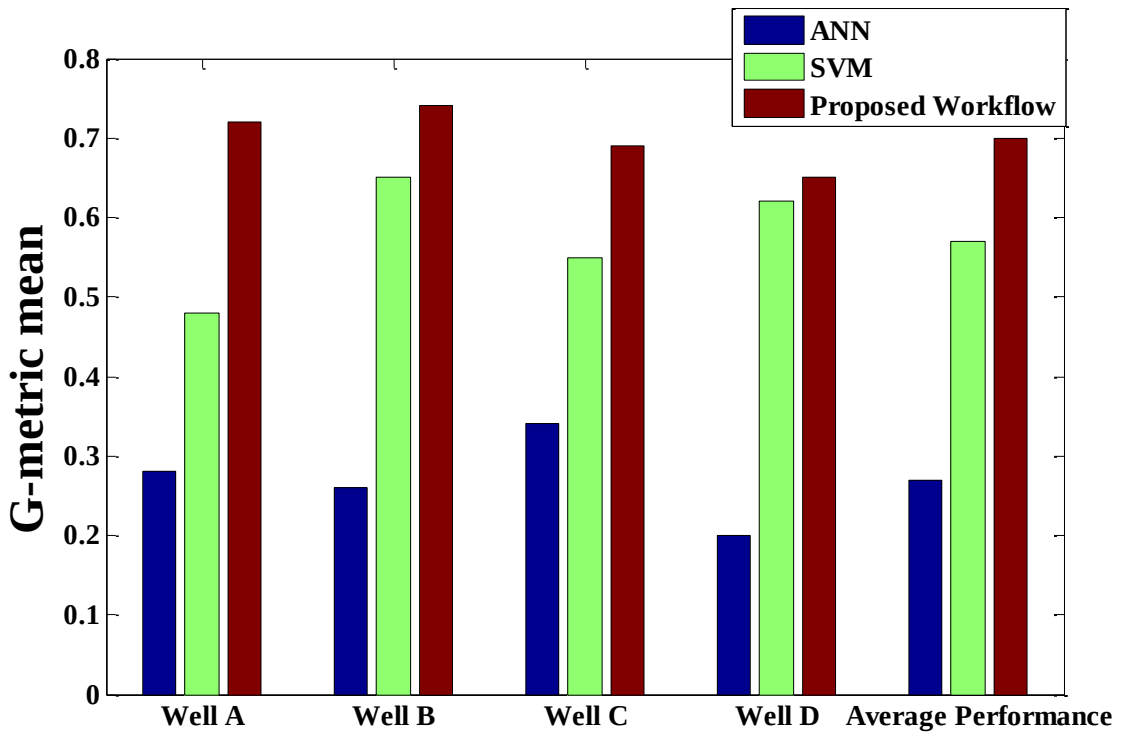

Fig. 5 : Bar plot describing performance of classifiers in terms of g-metric means

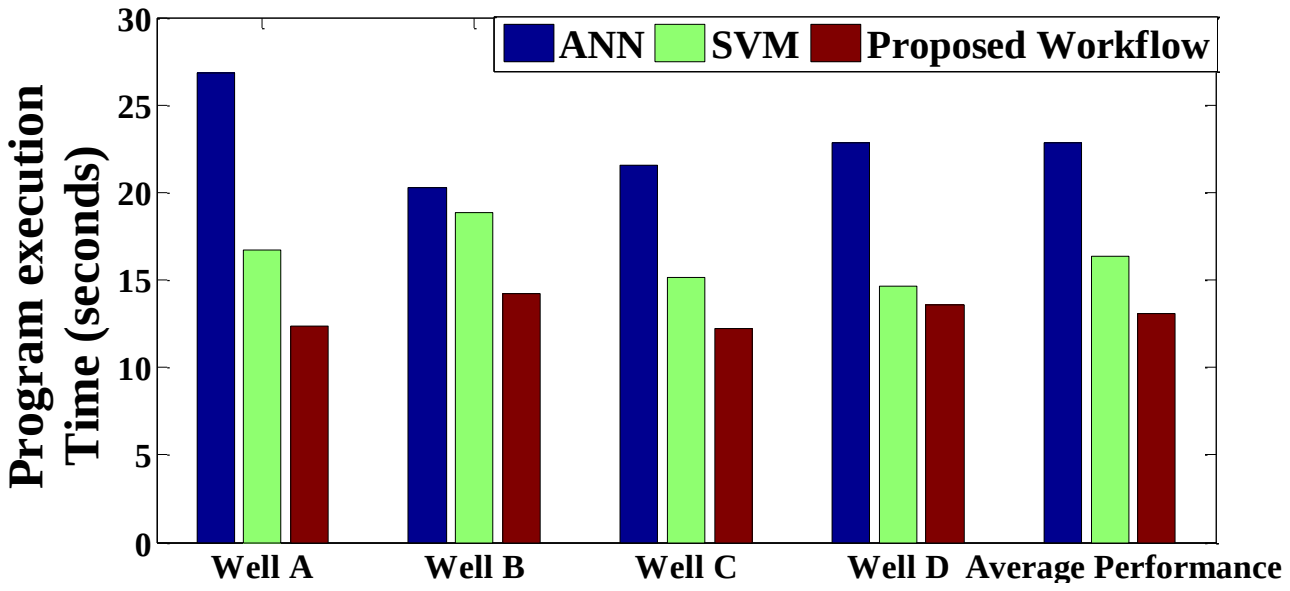

Fig. 6 :Bar plot describing performance of classifiers in terms of program execution time (in seconds)

### D. *Volumetric Classification and Visualization*

The proposed framework is established as a powerful tool to classify reservoir characteristics from seismic attributes. Fig.

7 represents the variation of seismic impedance, at a particular inline over the study area.

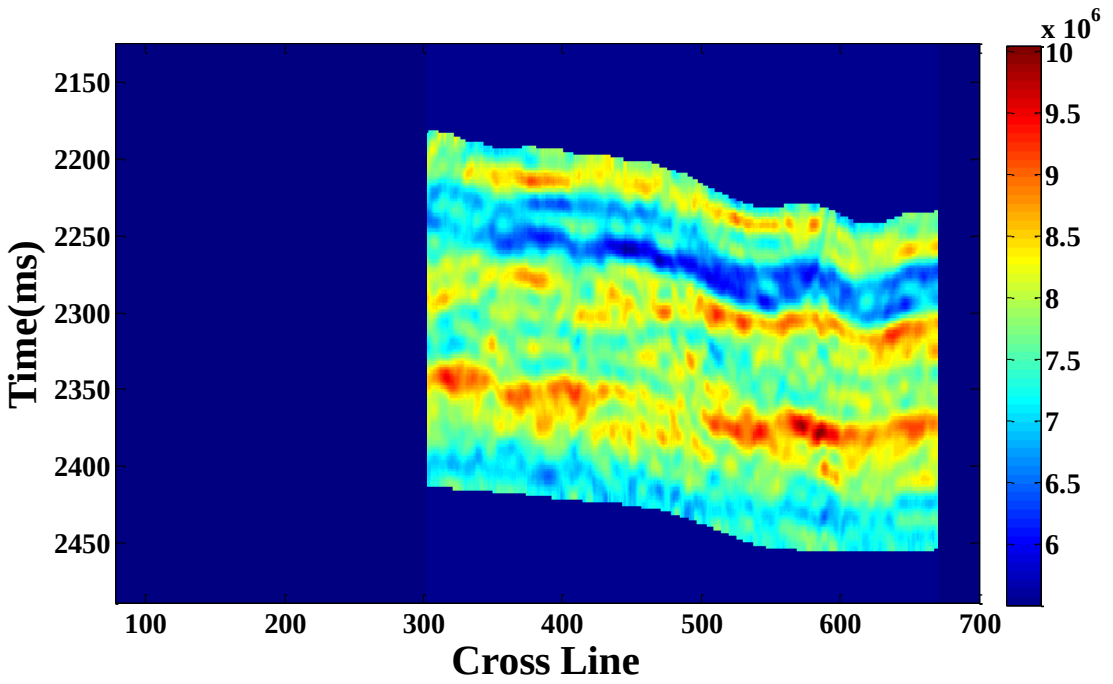

Fig. 7: Seismic impedance variation at inline 159

The tuned classifier which was saved while blind prediction of well A is further used to classify water saturation over the study area from predictor seismic signals.

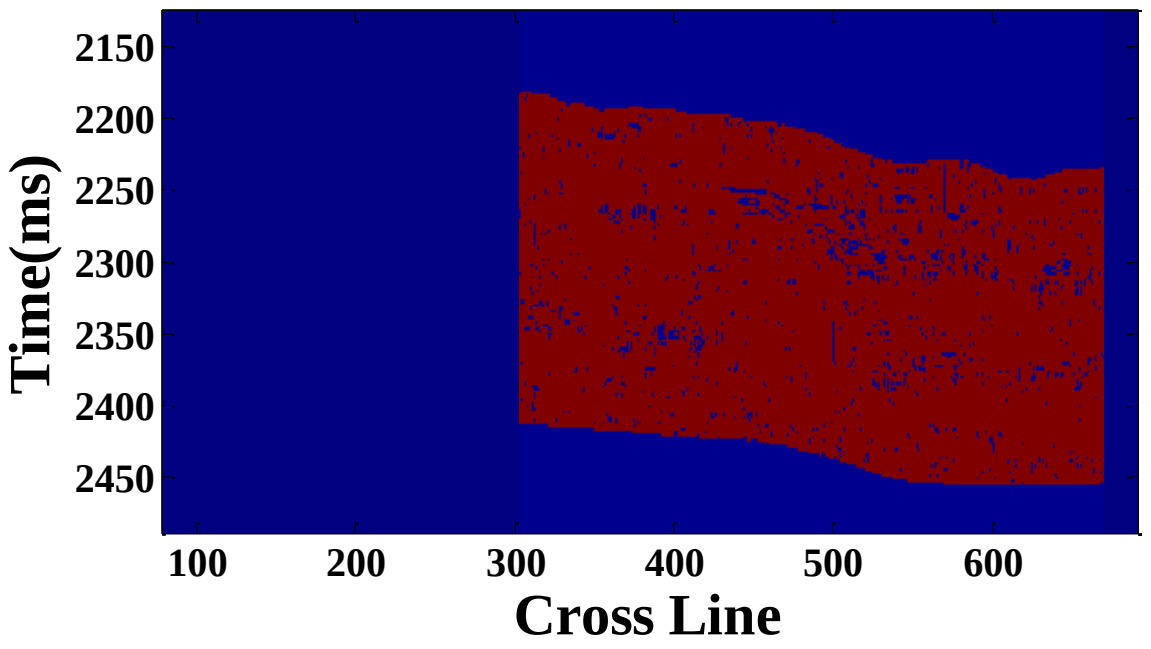

Fig. 8: Water saturation level variation at inline 159 (Red: Class high; Blue: Class low)

Fig. 8 represents distribution of water saturation level classified in two categories: Class high and Class low over the area at the same inline. Inside the study area, blue represents Class low and red color represents Class high samples. It can be observed from Fig. 8 that the presence of Class high patterns is significant over that of the Class low samples throughout the area.

## VI. Conclusion and Future Scope

This paper has proposed a classification framework to classify water saturation levels from seismic attributes using a small imbalanced dataset. Application of SVDD to solve a classification problem integrating seismic and well log signals in reservoir characterization field is a contribution of this paper. The area map representing high and low water saturation level is created using the proposed framework. Although the framework has outperformed existing supervised classifiers in terms of performance evaluator, there is a scope of improvement in selection of parameters associated with SVDD algorithm. The selection of SVDD parameters are carried out empirically keeping the improvement in classification in view. In future, efforts can be made to automate the selection procedure using some evolutionary algorithms such as genetic algorithm, particle swarm optimization etc.